\pdfoutput=1

\documentclass[11pt]{article}

\usepackage[table]{xcolor}

\usepackage[table]{xcolor}

\usepackage[]{emnlp2023}
\usepackage{times}
\usepackage{latexsym}
\usepackage{graphicx}

\usepackage[T1]{fontenc}
\usepackage{booktabs}
\usepackage{multirow}
\usepackage{colortbl}

\usepackage[utf8]{inputenc}

\usepackage{microtype}

\usepackage{gb4e}
\usepackage{amsmath}
\usepackage{centernot}
\usepackage{enumitem}

\noautomath

\title{\textsc{ProPres}: Investigating the Projectivity of Presupposition\\ with Various Triggers and Environments}

\author{Daiki Asami \\
  University of Delaware \\ %/ Address line 1 
  \texttt{daiasami@udel.edu} \\\And
  Saku Sugawara \\
  National Institute of Informatics \\ % / Address line 1 \\
  \texttt{saku@nii.ac.jp} \\}

\begin{document}
\maketitle
\begin{abstract}
What makes a presupposition of an utterance---information taken for granted by its speaker---different from other pragmatic inferences such as an entailment is projectivity (e.g., the negative sentence \textit{the boy did not stop shedding tears} presupposes \textit{the boy had shed tears before}).
The projectivity may vary depending on the combination of presupposition triggers and environments. However, prior natural language understanding studies fail to take it into account as they either use no human baseline or include only negation as an entailment-canceling environment to evaluate models' performance.
The current study attempts to reconcile these issues.
We introduce a new dataset, projectivity of presupposition (\textsc{ProPres}), which includes 12k premise--hypothesis pairs crossing six triggers involving some lexical variety with five environments.
Our human evaluation reveals that humans exhibit variable projectivity in some cases.
However, the model evaluation shows that the best-performed model, DeBERTa, does not fully capture it.
Our findings suggest that probing studies on pragmatic inferences should take extra care of the human judgment variability and the combination of linguistic items.
\end{abstract}

\section{Introduction}
It is an open question as to whether language models can learn a human-like pragmatic inference \cite{pavlick2022semantic}. 
A speaker does not always explicitly say everything in an utterance, but a hearer can infer what is implicit in it.
One notable case concerns a \textit{presupposition} that refers to information taken for granted by a speaker of an utterance \cite{stalnaker:74,beaver:97}.
Presuppositions are prevalent in our everyday communication; hence, a comprehensive investigation of whether models can understand them in the same way as humans can contribute to the development of a better language understanding system.

\begin{figure}[t]
    \centering
    \includegraphics[trim=0 0 0 0.1cm,clip, width=\linewidth]{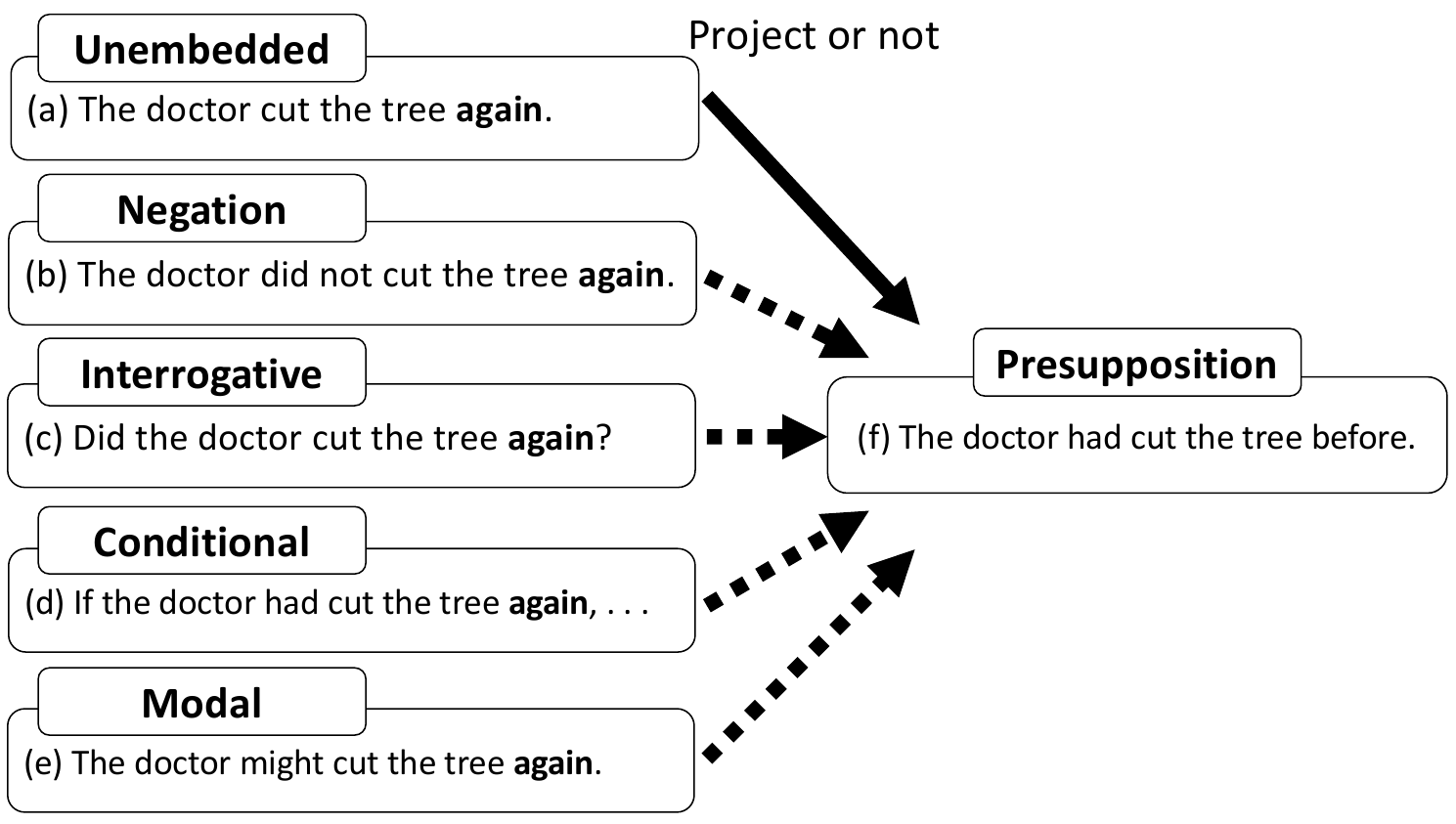}
    \caption{Projectivity of presupposition. A presupposition can project out of entailment-canceling environments. The dashed arrows indicate that the projectivity varies depending on the combination of triggers and environments.}
    \label{fig.1}
\end{figure}

Presupposition triggers introduce presuppositions (e.g., \textit{again} in Figure~\ref{fig.1} (a)).
A presupposition of (a) is \textit{the doctor had cut the tree before} (f).
What makes the presupposition different from an entailment (in this case, \textit{the doctor cut the tree one more time}) is projectivity:
the presupposition projects out of entailment-canceling environments (e.g., negative (b), interrogative (c), conditional (d), and modal (e) sentences) while the entailment does not.\footnote{Formal semantic and pragmatic literature generally uses the term, operators, rather than environments but we use the latter for the sake of readability.}
In other words, the presupposition (f) holds in the environments (b--e), but the entailment (\textit{the doctor cut the tree one more time}) does not.

\begin{table*}
\small
\centering \def\arraystretch{1.2}
\setlength{\tabcolsep}{5pt}
\rowcolors{1}{white}{gray!25}
\begin{tabular}{cccc} \toprule
Trigger Type & Example Triggers & Example Premise\\ \midrule
Iterative & \textit{again} & The assistant split the log \textbf{again}. \\
Aspectual verb & \textit{stop}, \textit{quit}, \textit{finish} & The assistant \textbf{stopped} splitting the log. \\
Manner adverb &  \textit{quietly}, \textit{slowly}, \textit{angrily} & The assistant split the log \textbf{quietly}. \\
Factive verb & \textit{remember}, \textit{regret}, \textit{forget} & The assistant \textbf{remembered} splitting the log. \\
Comparative & \textit{better than}, \textit{earlier than} & The assistant split the log \textbf{better than} the girl. \\
Temporal adverb &\textit{before}, \textit{after}, \textit{while} & The assistant split the log \textbf{before} bursting into the room.
\\ \bottomrule
\end{tabular}
\caption{Presupposition triggers with an affirmative (unembedded) premise in \textsc{ProPres}.}
\label{unembedded triggers}
\end{table*}

\begin{table*}
\small
\centering \def\arraystretch{1.2}
\setlength{\tabcolsep}{4pt}
\begin{tabular}{cccc} \toprule
Environment & Premise & Hypothesis (target and control) & Label (target/control)\\ \midrule
Unembedded & The doctor shed tears again. & \multirow{2.5}{*}{Target: The doctor had (not) shed tears before.} & E (C) / E (C)\\
Negation & The doctor did not shed tears again. & & E (C) / C (E)\\
Interrogative & Did the doctor shed tears again? & \multirow{3}{*}{Control: The doctor (did not) shed tears again.} & E (C) / N (N)  \\
Conditional & If the doctor had shed tears again, ... & & E (C) / C (E)\\
Modal & The doctor might shed tears again. & & E (C) / N (N) \\ \bottomrule
\end{tabular}
\caption{
    Environments used in \textsc{ProPres}.
    \textit{E} = \textit{Entailment}, \textit{C} = \textit{Contradiction}, and \textit{N} = \textit{Neutral}.
    The labels in the target conditions are defined based on projectivity.
    }
\label{premise hypothesis pair}
\end{table*}

Crucially, linguistic studies suggest that the projectivity can vary depending on many factors \cite{karttunen:71,simons:01,sevegnani-etal-2021-otters,tonhauser:18,tonhauser2019information,degen2021a}.
Previous probing studies in natural language processing examine models' performance on presuppositions in the natural language inference (NLI) task \cite{jeretic-etal-2020-natural, parrish-etal-2021-nope}.
However, they do not fully take into account the variable aspect of the projectivity.
For instance, \citet{jeretic-etal-2020-natural} obtain no human baseline, which makes models' performance hard to interpret.
\citet{parrish-etal-2021-nope} collect human data but use only one entailment-canceling environment, negation.
Hence, it remains unclear about the projectivity out of other environments.

This study attempts to reconcile these issues.
We first evaluate recent pretrained language models against a presupposition portion of \textsc{ImpPres} \citep{jeretic-etal-2020-natural}.
Specifically, we conduct a human evaluation on its subset (900 pairs), each of which ends up receiving 9.4 labels on average, and then evaluate RoBERTa~\cite{liu2019roberta} and DeBERTa~\cite{he2020deberta}.
We find that humans exhibit relatively weak projectivity in some examples but the best-performed model, DeBERTa, does not perform in a human-like way.

\textsc{ImpPres} is imperfect in terms of comprehensiveness: the nine triggers that it uses are not exhaustive (cf. \citet{Levinson:83} and \citet{Potts:15} list a total of 27 triggers) and are lexically limited.
Thus, using six new triggers with some lexical variety (Table~\ref{unembedded triggers}) and five environments (Table~\ref{premise hypothesis pair}), we construct an extensive evaluation dataset: projectivity of presupposition (\textsc{ProPres}), which consists of 12,000 sentence pairs.
We evaluate four models (bag-of-words,
InferSent~\cite{conneau-etal-2017-supervised}, RoBERTa, and DeBERTa) with \textsc{ProPres} against human judgments on its subset (600 pairs)
Each pair has more than 50 human labels on average.
This second evaluation reveals that human data exhibit variable projectivity not only in previously attested cases such as manner adverbs in interrogative and negative environments \cite{stevens2017rational,tonhauser:18} but also in unattested cases such as those in conditional and modal environments.
Additionally, we find some within-trigger-type variation.
However, the best-performed model, DeBERTa, shows poor performance on controls and does not fully capture the variable projectivity patterns, indicating that it does not learn the pragmatic knowledge necessary to understand presuppositions.
These findings suggest that the combination of the various linguistic items in \textsc{ProPres} and the human evaluation allow us to probe the model's behavior more adequately.

The results from our two evaluations suggest that studies evaluating language understanding systems and creating datasets targeting pragmatic inferences should take extra care of the human judgment variability and the combination of linguistic items.
In conclusion, this study makes the following contributions:\footnote{Our dataset with the human labels and codes used to generate it are available at \url{https://github.com/nii-cl/projectivity-of-presupposition}.}
\begin{itemize}[leftmargin=4mm]
    \item We introduce \textsc{ProPres} using six novel presupposition triggers embedded under five environments, which enables a comprehensive investigation of the projectivity of presupposition.
    \item Our human evaluation provides evidence for the variable projectivity depending on the combination of triggers and environments.
    \item Our model evaluation against human results reveals that the models and humans behave differently in the understanding of presuppositions.
\end{itemize}

\section{Background}
\subsection{Presupposition in Linguistics}
\label{Presupposition in Linguistics}
Linguistic items or constructions introducing a presupposition are referred to as presupposition triggers \cite[e.g., \textit{again} in Figure~\ref{fig.1};][]{stalnaker:74,beaver:97}.
One property that makes presuppositions distinct from other pragmatic inferences such as an entailment is projectivity: presuppositions survive in entailment-canceling environments such as negation \cite{karttunen:73, heim:83}.
For instance, a presupposition of the affirmative sentence with the presupposition trigger \textit{again} ((f) given (a)) holds when embedded under negation (b).
In contrast, the same environment cancels an entailment (here, \textit{the doctor cut the tree one more time}).

Importantly, previous linguistic studies show that the projectivity of presupposition can vary depending on factors such as context, lexical items, prior beliefs, a speaker's social identity, and prosodic focus
\cite{karttunen:71,simons:01,stevens2017rational,tonhauser:18,tonhauser2019information, degen2021a}.
This variability is in line with the observation that humans make unsystematic judgments about projectivity on both natural \cite{ross-pavlick-2019-well, marneffe:19} and controlled \cite{white:18} sentences.
One remaining question here is whether the variable projectivity has to do with the interaction of triggers and environments (e.g., is a presupposition triggered by \textit{again} more likely to project over the negation (b) than the conditional (d)?).
To tackle this question comprehensively, this study collects human judgments on presuppositions using a wide range of triggers and environments.

\subsection{Presupposition in NLI}
Previous studies introduce NLI datasets to evaluate model performance on presuppositions \cite{jeretic-etal-2020-natural, parrish-etal-2021-nope}.
One example is a template-based dataset: \textsc{ImpPres} \cite{jeretic-etal-2020-natural}.
Using this dataset, \citet{jeretic-etal-2020-natural} conclude that models (e.g., BERT \cite{devlin-etal-2019-bert}) learn the projectivity of presuppositions triggered by \textit{only}, cleft existence, possessive existence, and question.
However, there is one problem with them, that is, no human evaluation.
As discussed in Section \ref{Presupposition in Linguistics}, it is possible that projectivity varies depending on the combination of triggers and environments.
Thus, it is unknown whether the results of the model evaluation reported by \citet{jeretic-etal-2020-natural} align with human data.
To solve this issue, following \citet{parrish-etal-2021-nope}, we conduct human evaluation on a subset of \textsc{ImpPres} as well as our dataset, \textsc{ProPres}.

Another dataset relevant to our study is NOPE \cite{parrish-etal-2021-nope}, which consists of naturally-occurring sentences with presupposition triggers.
With this dataset, \citet{parrish-etal-2021-nope} evaluate transformer-based models against human performance, finding that models behave similarly to humans.
One limitation of NOPE is that it includes only negation as an entailment-canceling environment.
As a result, the generalizability of the findings by \citet{parrish-etal-2021-nope} is unclear beyond negation.
To draw a more general conclusion, it is necessary to include various types of environments.
Following \citet{jeretic-etal-2020-natural}, the entailment-canceling environments in \textsc{ProPres}, include not only negation but also an interrogative, conditional, and modal.

\section{Experiment 1: Reevaluating \textsc{ImpPres}}
One limitation in \citet{jeretic-etal-2020-natural} is no human evaluation, which leaves it open whether models capture any variable projectivity exhibited by humans.
To overcome it, we collect human labels on a subset of \textsc{ImpPres}, testing the performance of the two models, RoBERTa and DeBERTa, against the human results.

\subsection{Setup} \label{experiment 1 setup}
\paragraph{Human Evaluation}
Our human evaluation targets a subset of \textsc{ImpPres}, which uses nine triggers (\textit{all N}, \textit{both}, change of state verbs (CoS), cleft existence, \textit{only}, possessive definites, possessive uniqueness, and question).
Specifically, we focus on conditions where triggers occur in one of the five environments (the affirmative sentence (unembedded), negative sentence (negation), conditional antecedent (conditional), modal sentence (modal), and interrogative)\footnote{Examples of triggers and environments in \textsc{ImpPres} appear in Appendix~\ref{appendix triggers and environments in imppres}.}
and where a hypothesis is either an affirmative or negative sentence.
We randomly extract ten items from each condition (a total of 900 sentences).

Using Amazon Mechanical Turk,\footnote{\url{https://www.mturk.com}} we conduct the human evaluation run on PCIbex.\footnote{\url{https://farm.pcibex.net}}
Figure~\ref{pcibex} shows an example prompt that we use in the human evaluation.
%During the experiment, the following instruction appears on a screen: \textit{Select the response based on how likely you think the second statement is to be true, using the information in the first statement and your background knowledge about how the world works.
%If you think that the second statement is true, click Entailment.
%If you think that it is false, select Contradiction.
%If you are not sure, select Neutral}.
We adopt and modify the instruction for the human evaluation from \citet{parrish-etal-2021-nope}.
As a result of the human evaluation, each of the extracted items receives 9.4 labels on average.\footnote{Appendix~\ref{qualification} reports more details of the human evaluation (e.g., crowdsourcing qualification and exclusion criteria).}

\paragraph{Model Evaluation}
We evaluate Huggingface’s \cite{wolf-etal-2020-transformers} pretrained RoBERTa-base \cite{liu2019roberta} and DeBERTa-v3-large \cite{he2020deberta} fine-tuned on MNLI \cite{williams-etal-2018-broad}.
We do not evaluate a bag-of-words (BOW) model and an InferSent model \cite{conneau-etal-2017-supervised} because \citet{jeretic-etal-2020-natural} show that their accuracy for control conditions is below chance (33.3\%).

\begin{figure}[t]
    \centering
    \includegraphics[trim=0 0 0 0.1cm,clip, width=\linewidth]{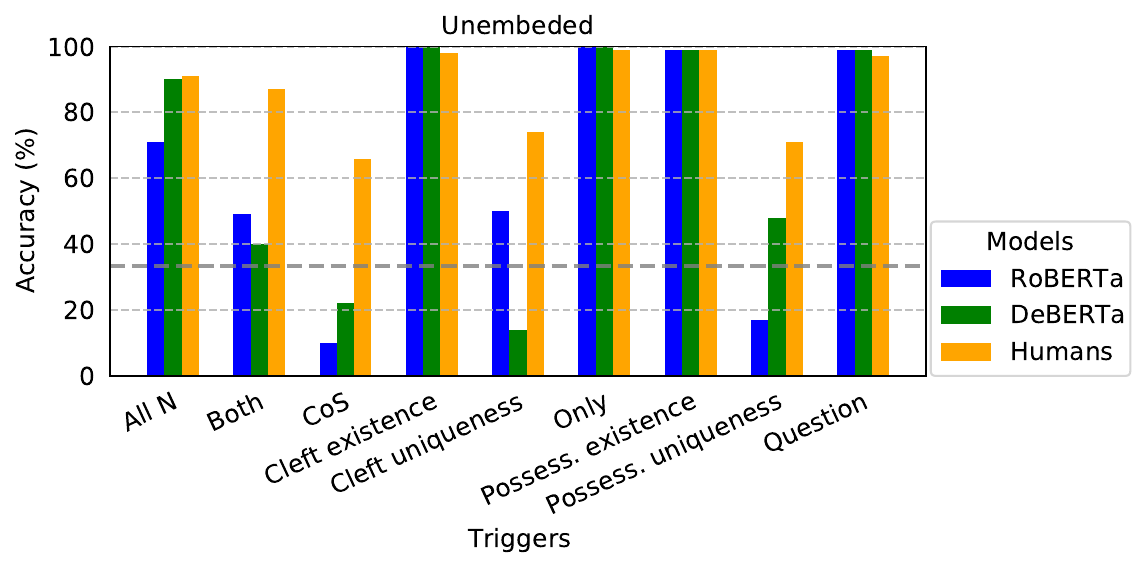} %, height=4.5cm]{unembedded.pdf}
    \caption{Results on the unembedded triggers in \textsc{ImpPres}.
    The dashed lines indicate chance performance (33.3\%).}
    \label{unembedded}
\end{figure}

\begin{figure}[t]
    \centering
    \includegraphics[trim=0 0 0.0 0,clip,width=\linewidth]{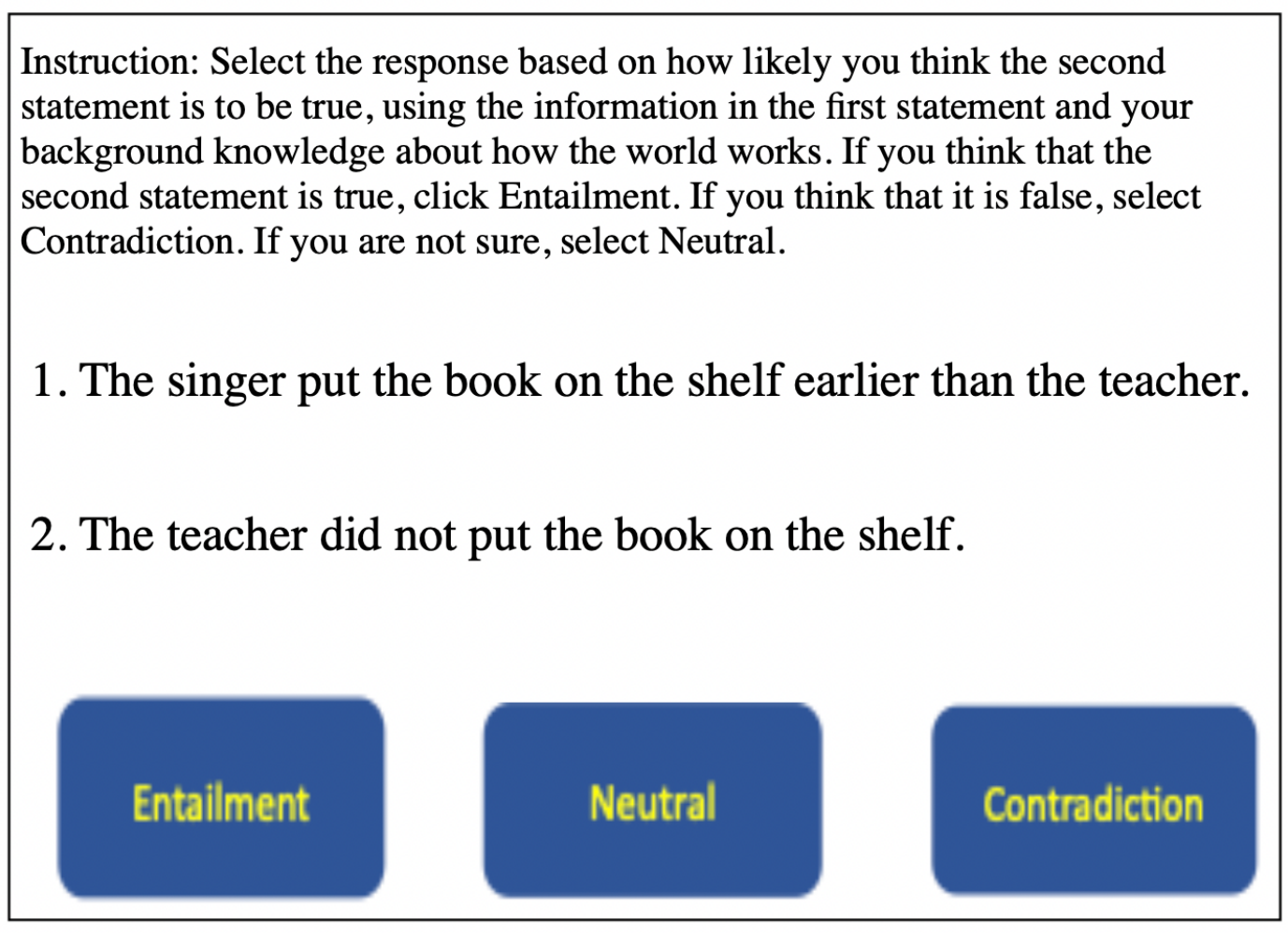}
    \caption{An example prompt in the human evaluation.}
    \label{pcibex}
\end{figure}

\subsection{Results and Discussion}\label{ex1 results and discussion}
\paragraph{Unembedded Triggers}
We use accuracy for the unembedded triggers as criteria to exclude triggers from the analysis of entailment-canceling environments.
When a trigger occurs in an affirmative sentence (unembedded), a presupposition equals an entailment  (e.g., \textit{Bob only ran} presupposes and entails \textit{Bob ran}) \cite{jeretic-etal-2020-natural}.
If humans show low accuracy for any unembedded triggers, we manually analyze the relevant triggers to identify their cause.
We interpret models' low accuracy as lack of knowledge of relevant triggers if humans show high accuracy for the same triggers.

The results of the human evaluation (Figure~\ref{unembedded}) show lower accuracy for CoS (66.3\%), cleft uniqueness (74.1\%), and possessed uniqueness (71.9\%), examples of which are provided below, compared to the other triggers (acc. $>$ 87.3\%).\footnote{Throughout the paper, the examples from the dataset are slightly simplified (e.g., changing \textit{Thomas} to \textit{Tom}) for the space reason.}

\begin{exe}
\ex \label{cos}
CoS: Omar is hiding Ben.\\
$\rightarrow$ Ben was out in the open.
\end{exe}

\begin{exe}
\ex \label{cleft}
Cleft uniqueness: It is that doctor who left.\\
$\centernot \rightarrow$ More than one person left.
\end{exe}

\begin{exe}
\ex \label{poss}
Possessive uniqueness: Tom's car that broke bored this committee.\\
$\rightarrow$ Tom has exactly one car that broke.
\end{exe}

We reason that the low accuracy for CoS is due to lexical ambiguity.
For instance, people might label the pair (\ref{cos}) as neutral or contradiction because Ben was not necessarily exposed before being hidden.
Regarding the other two conditions, we do not understand the exact source of the low accuracy at this point.
In linguistics, results from human judgment experiments sometimes contradict generalizations made by theoreticians \citep{gibson2013need}.
Additionally, NLI research reports disagreements in human labels \cite{pavlick-kwiatkowski-2019-inherent,nie-etal-2020-learn,zhang-de-marneffe-2021-identifying,jiang-marneffe-2022-investigating}.
Thus, the current results suggest that judgments on presuppositions of cleft and possessive uniqueness are not as robust as \citet{jeretic-etal-2020-natural} might assume.
Consequently, we remove CoS, cleft uniqueness, and possessed uniqueness from the following analysis as they might confound the results.

\begin{figure*}[t]
    \centering
    \includegraphics[trim=0 0 0 0.1cm,clip,width=0.9\textwidth]{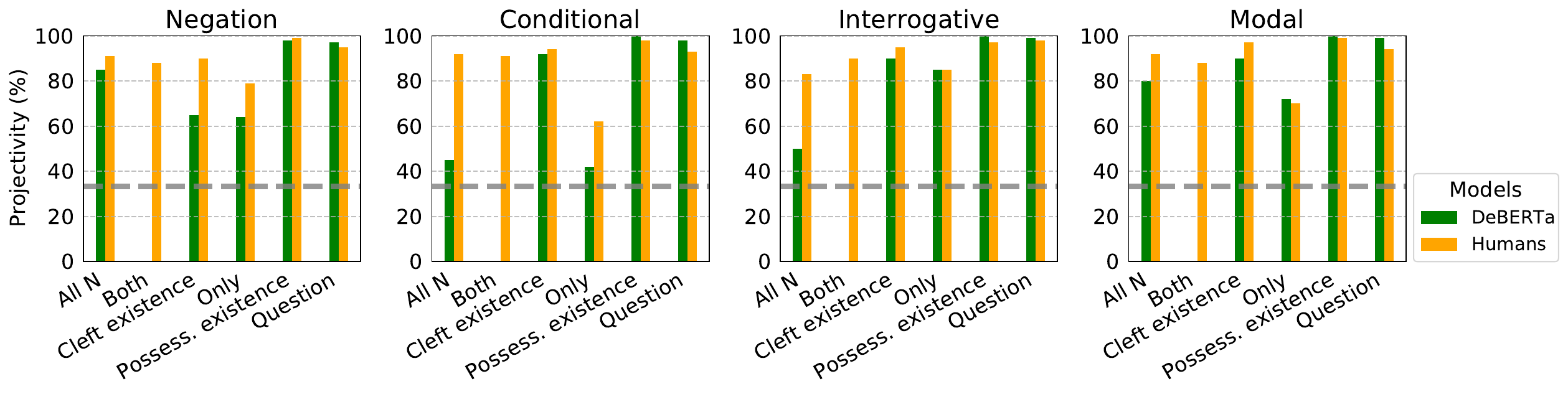}   %, height=3.5cm
    \caption{Results on entailment-canceling environments in \textsc{ImpPres}.
    DeBERTa's results on \textit{both} are not presented.}
    \label{imppres four conditions}
\end{figure*}

The results of the model evaluation reveal that both RoBERTa and DeBERTa achieve high accuracy for most triggers (acc. $>$ 89.5\%).
Two exceptions are \textit{all N} and \textit{both}.
RoBERTa shows lower accuracy for \textit{all N} (71.0\%) than DeBERTa (89.5\%) (e.g., \textit{all four men that departed telephoned} $\rightarrow$  \textit{exactly four men departed}).
With respect to \textit{both} (e.g., \textit{both guys who ran jumped} $\rightarrow$ \textit{exactly two guys ran}), neither DeBERTa nor RoBERTa performs well (39.0\% and 49.0\%, respectively).
Otherwise, the two models are roughly comparable in performance.
Thus, we analyze only DeBERTa.

Based on the human and model results, our analysis of entailment-canceling environments includes the five triggers: \textit{all N}, cleft existence, \textit{only}, possessive existence, and question.\footnote{We report all results including excluded triggers in Appendix~\ref{appendix results without exclusion}.}

\paragraph{Entailment-Canceling Environments}
To analyze results on entailment-canceling environments, we use the term, projectivity, instead of accuracy.
Since human judgments on projectivity can vary, as discussed in Section~\ref{Presupposition in Linguistics}, we should not define gold labels for sentence pairs involving presupposition.
We calculate projectivity based on whether presupposition holds when embedded under an entailment-canceling environment.
For instance, if one classifies the pair, \textit{did Tom only terrify Ken?} and \textit{Tom terrified Ken}, as entailment, we consider it as projective.
Taking another example, if one judges the hypothesis \textit{Tom did not terrify Ken} as contradiction given the same premise, it counts as projective.
Otherwise, we take these two examples as non-projective.

Figure~\ref{imppres four conditions} presents results on the four environments: negation, conditional, interrogative, and modal.
Overall, DeBERTa and humans behave similarly.
For instance, they show relatively low projectivity in \textit{only} in conditional (e.g., \textit{if Mary only testifies, ...} $\rightarrow$ \textit{Mary testifies}) and modal (e.g., \textit{Mary might only testify} $\rightarrow$ \textit{Mary testifies}) (61.8\% and 69.8\% for humans and 41.5\% and 72.0\% for DeBERTa, respectively).

A closer look at the results reveals that DeBERTa takes some conditions less projective than humans.
Humans take cleft existence in negation (e.g., \textit{it isn't that guest who complained} $\rightarrow$ \textit{someone complained}) as projective (89.7\%) while DeBERTa predicts it as less projective (65.0\%).
In addition, humans judge \textit{all N} in conditional (e.g., \textit{if all nine actors that left slept, ...} $\rightarrow$ \textit{exactly nine actors left}) and in interrogative (e.g., \textit{did all nine actors that left sleep?} $\rightarrow$ \textit{exactly nine actors left}) as projective (91.8\% and 82.6\%, respectively) but DeBERTa takes them as less projective (45.0\% and 49.5\%, respectively).
These results indicate DeBERTa's lack of knowledge of cleft existence in negation and \textit{all N} in conditional and interrogative.

In summary, humans take most presupposition cases as projective except \textit{only} embedded under conditional and modal.
This finding adds to the previous research on variable projectivity in other cases \cite{stevens-guille-etal-2020-neural,tonhauser:18,tonhauser2019information, degen2021b, degen2021a}.
Additionally, DeBERTa and humans show not only similarities but also differences in projectivity.

\section{Experiment 2: \textsc{ProPres}}
An investigation of the projectivity of presupposition with \textsc{ImpPres} is far from comprehensive because we can find more triggers in the literature (e.g., 27 triggers in \citet{Levinson:83} and \citet{Potts:15} in total) and none of the six triggers which we analyze in \textsc{ImpPres} has lexical variation.
Using six additional triggers with some lexical variety, we create a new dataset, \textsc{ProPres}, which allows us to investigate the variable projectivity and models' behavior more comprehensively.

\subsection{Data Generation}
\paragraph{Triggers and Environments}\label{paragraph triggers and environments}
\textsc{ProPres} has six types of presupposition triggers: (1) the iterative \textit{again}, (2) aspectual verbs, (3) manner adverbs, (4) factive verbs, (5) comparatives, and (6) temporal adverbs, as presented in Table~\ref{unembedded triggers}.
We select these triggers from \citet{Levinson:83} and \citet{Potts:15}  because they are not included in \textsc{ImpPres} and can be easily incorporated into templates.
Crucially, these triggers allow us to use different lexical items (e.g., we use seven verbs and nine adverbs for aspectual verbs and manner adverbs, respectively).
One exception is \textit{again}, but it is a standard presupposition trigger investigated by theoretical linguistic \cite{vonStechow1995lexical,bale2007quantifiers} and natural language processing \cite{cianflone-etal-2018-lets} research.
Thus, it is worth including this trigger in the dataset.

\textsc{ProPres} uses five environments: (1) affirmative sentences (unembedded), (2) negative sentences (negation), (3) polar questions (interrogative), (4) counterfactual conditional antecedents (conditional), and (5) modal sentences (modal), as exemplified in Table~\ref{premise hypothesis pair}.
We include the unembedded environment to test whether humans and models can identify presupposition as entailment when triggers occur in affirmative sentences.
The counterfactual conditional antecedent is not a typical entailment-canceling environment, but we include it to ensure that conditional controls have clear gold labels (entailment or contradiction) as we discuss in the following paragraph.
We generate affirmative and negative hypotheses for each premise sentence.
Combining six trigger types, five environment types, and two hypothesis polarity types results in 60 conditions.
Generating 100 premise--hypothesis pairs for each condition yields 6,000 pairs.\footnote{We provide examples for each condition in Appendix~\ref{appendix templates}.}

We make a control condition corresponding to each target condition where a hypothesis is either an affirmative or negative version of its premise, as shown in Table~\ref{premise hypothesis pair}.
The control conditions serve as a sanity check in a human evaluation.
They are also important to test whether the models rely on lexical overlap \cite{mccoy-etal-2019-right} or negation \cite{gururangan-etal-2018-annotation} heuristics.
For instance, models should label the affirmative hypothesis in Table~\ref{premise hypothesis pair} as entailment if they rely on the lexical overlap heuristic because of the high lexical overlap between the premise and hypothesis.
Additionally, they should label the negative hypothesis with \textit{not} as contradiction if they use the negation heuristic.
Only if models predict correctly in the control conditions, we can say that their predictions about the corresponding target conditions indicate projectivity rather than heuristics.
Creating 100 pairs for each control condition results in 6,000 pairs. 
In total, \textsc{ProPres} comprises 12,000 sentence pairs.

\paragraph{Templates}
We make templates and generate sentences with them using the codebase developed by \citet{yanaka-mineshima-2021-assessing}.\footnote{\url{https://github.com/verypluming/JaNLI}}
Following are examples of templates and sentences.\footnote{A full list of the templates and their example sentences appears in Appendix~\ref{appendix templates}.}

\begin{exe}
\ex
The N did not VP again.\\
(The girl did not hurt others again.)\\
$\rightarrow$ ($\centernot \rightarrow$) The N had (not) VP before.\\
{} (The girl had (not) hurt others before.)
\end{exe}

\noindent In VP, we use verbs having the same form in past tense and past participle forms (e.g., \textit{hurt}) to make the morphological difference between a premise and hypothesis as small as possible.
This is crucial to check whether models rely on the lexical overlap heuristic in the control conditions.

\begin{figure*}[t]
    \centering
    \includegraphics[trim=0 0 0 0.1cm,clip,width=0.9\textwidth]{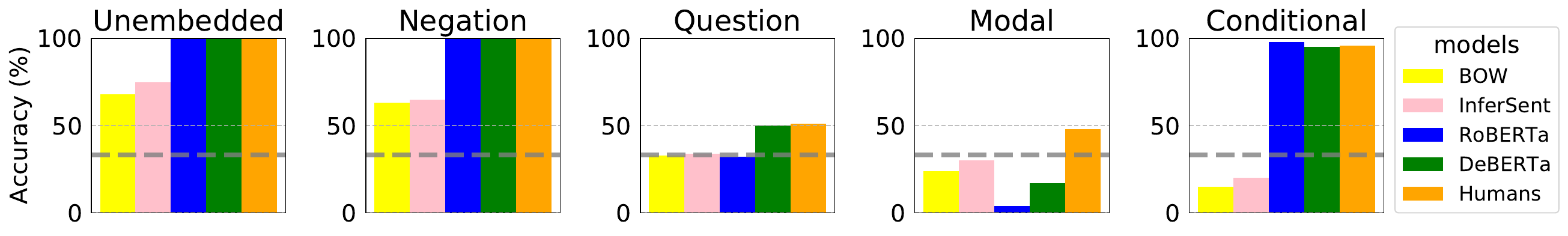} %,height=2.5cm
    \caption{Results on control conditions in \textsc{ProPres}.
    }
    \label{modelcontrol}
\end{figure*}

The use of templates has three advantages.
First, it allows us to systematically test whether models rely on the lexical overlap \cite{mccoy-etal-2019-right} and negation \cite{gururangan-etal-2018-annotation} heuristics.
In addition, it enables us to conduct a targeted evaluation with a large number of sentences including presupposition triggers embedded under particular environments.
Preparing the same number of data might be impossible if we use corpora.
Finally, we can rule out the effect of plausibility.
Previous linguistic work shows that the projectivity of presupposition varies depending on its content \cite{karttunen:71,simons:01,tonhauser:18}.
For instance, the sentence \textit{John didn't stop going to the restaurant} leads to the inference \textit{John had been going to the restaurant before}.
In contrast, the sentence \textit{John didn't stop going to the moon} is less likely to yield the inference \textit{John had been going to the moon before}.
This difference might stem from our world knowledge: it is more plausible for one to go to a restaurant than the moon.
As the plausibility effect is not the focus of this study, we use templates to control it.

\subsection{Setup}
\paragraph{Human Evaluation}
We randomly select ten out of 100 pairs from each target condition and two pairs from each control condition, extracting 600 and 120 pairs in total, respectively.
The human evaluation procedure is identical to the one reported in Section~\ref{experiment 1 setup}: using Amazon Mechanical Turk, we conduct the evaluation run on PCIbex.
As a result, each of the extracted pairs has 56.7 labels on average.
Due to some revision of \textsc{ProPres} during the dataset creation, we collect judgments on the modal environment and comparative trigger in Experiment 1 (200 pairs in total).
As a consequence, they receive 9.4 labels on average.

\paragraph{Model Evaluation}
We evaluate four models: BOW, InferSent \cite{conneau-etal-2017-supervised}, RoBERTa-base \cite{liu2019roberta}, and DeBERTa-v3-large \cite{he2020deberta}.
For the first two models, we follow \citet{parrish-etal-2021-nope}'s implementation\footnote{\url{https://github.com/nyu-mll/nope}} and use MNLI \cite{williams-etal-2018-broad} to fine-tune the parameters.
We use the GloVe embeddings for the word-level representations \cite{pennington-etal-2014-glove}.
For the two transformer-based models, we use RoBERTa-base and DeBERTa-v3-large fine-tuned on MNLI as in Experiment 1.

\begin{figure}[t]
    \centering
    \includegraphics[trim=0 0 0 0.1cm,clip, width=0.9\linewidth]{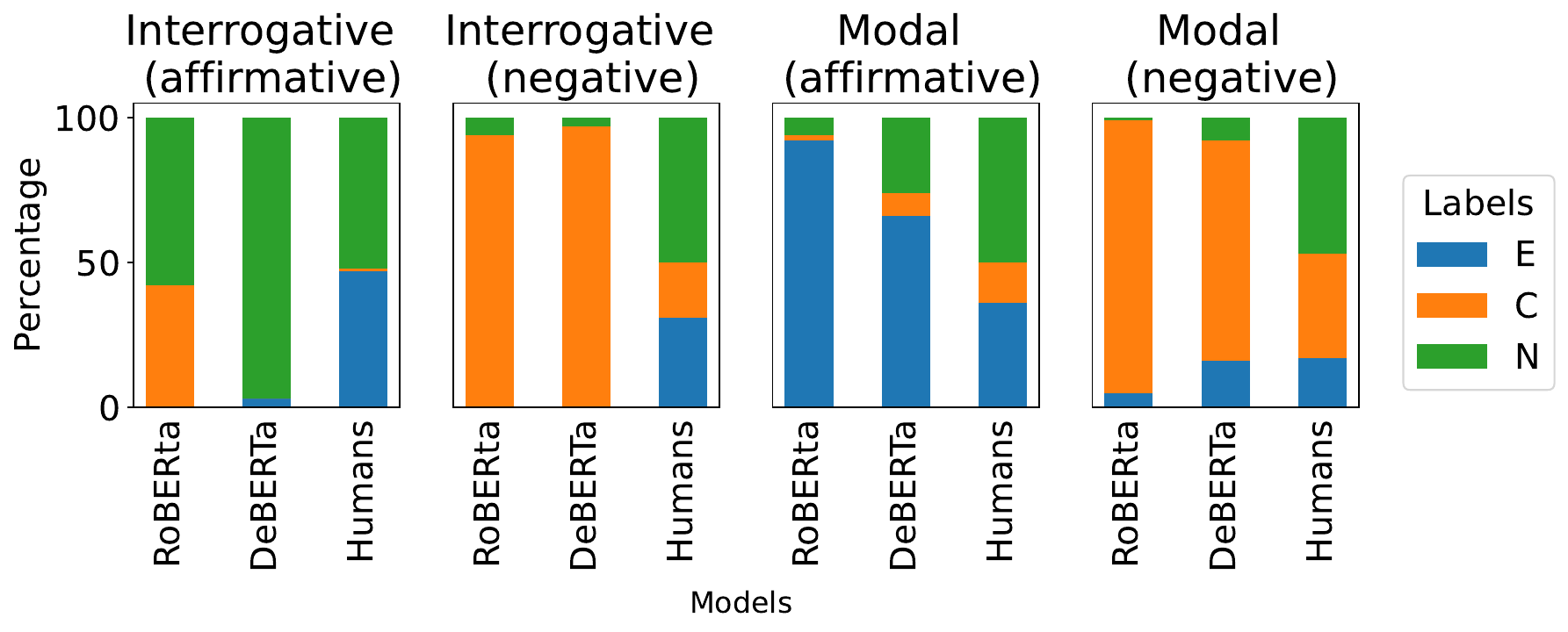} %  height=3.5cm
    \caption{Distributions of labels in the interrogative and modal with an affirmative or negative hypothesis.}
    \label{control selected}
\end{figure}

\subsection{Results and Discussion}
\paragraph{Control Conditions}
Figure \ref{modelcontrol} shows results on control conditions in which a hypothesis is either an affirmative or negative version of its premise.
The performance of InferSent and BOW models is poor, which makes their performance on target conditions hard to analyze.
Thus, we exclude them from our analysis below.
Similar to humans, RoBERTa and DeBERTa perform well on the unembedded, negation, and conditional (e.g., $P_1$--$P_3$ in (\ref{control examples in text})), indicating that they do not rely on the lexical overlap heuristic or \textit{negation} heuristic in these cases.

\begin{exe}
\ex 
\label{control examples in text}
    $P_1$: The boy cut the tree again.\\
    $P_2$: The boy did not cut the tree again.\\
    $P_3$: If the boy had cut the tree again, ... \\
    $P_4$: Did the boy cut the tree again?\\
    $P_5$: The boy might cut the tree again.\\ 
    $H_{1(2)}$: The boy (did not) cut the tree again.
\end{exe}

RoBERTa, DeBERTa, and humans perform poorly on the interrogative and modal (e.g., $P_4$ and $P_5$ in (\ref{control examples in text})) in which the correct label is supposed to be neutral \cite{jeretic-etal-2020-natural} (31.8\%, 50.0\%, and 51.1\% for interrogative and 3.5\%, 16.7\%, and 48.1\% for modal, respectively).
Distributions of labels in these conditions (Figure~\ref{control selected}) show that the majority of labels in humans are neutral, which is consistent with the view that a yes/no question does not have a truth value and thus one cannot decide whether its affirmative or negative version is true or not \cite{groenendijk1984,roberts2012information}.
One exception is the interrogative with an affirmative hypothesis (e.g., $P_4$ and $H_1$ in (\ref{control examples in text})): distributions of entailment and neutral are comparable (46.5\% and 52.4\%, respectively).
We suspect that some people interpret this condition as a confirmation question in which the affirmative counterpart of the interrogative (in this case, $H_1$) is presupposed, resulting in a high percentage of entailment.

In the same condition, the label distributions of DeBERTa and RoBERTa do not mirror those of humans.
RoBERTa shows a relatively high percentage of contradiction (57.5\%) whereas DeBERTa shows a very high percentage of neutral (97.1\%).
In the interrogative with the negative hypothesis (e.g., $P_4$ and $H_2$), RoBERTa and DeBERTa assign contradiction to the hypothesis the majority of the time (93.7\% and 97.1\%, respectively), indicating the negation heuristic: models are likely to label a given hypothesis as contradiction if it includes \textit{not} \cite{gururangan-etal-2018-annotation}.

\begin{figure*}[t]
    \centering
    \includegraphics[trim=0 0 0 0.1cm,clip,width=0.9\textwidth]{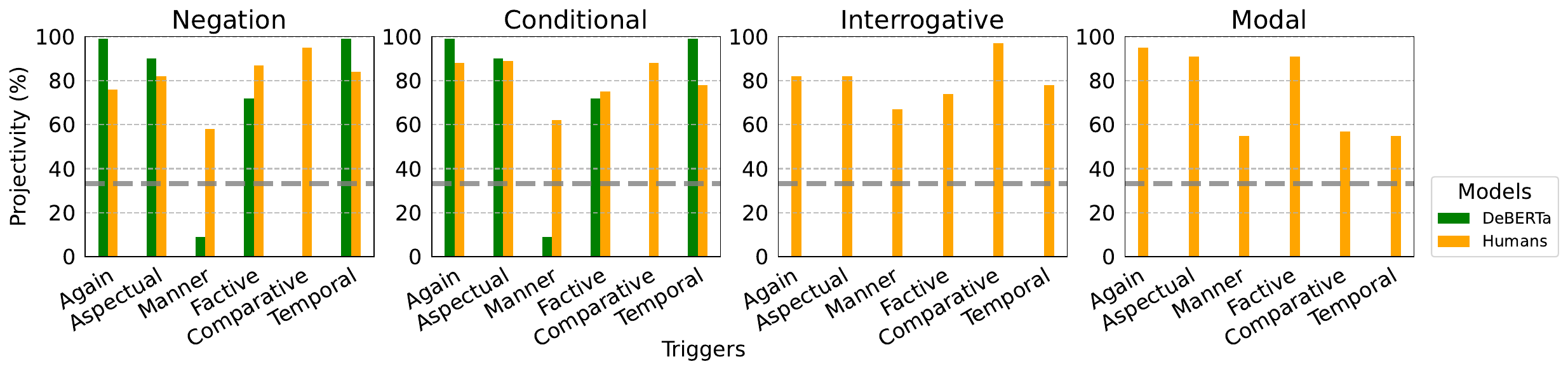} % , height=3.5cm
    \caption{Results on entailment-canceling environments in \textsc{ProPres}.
    DeBERTa's results on the interrogative and modal environments and the comparative trigger are not shown due to its unstable performance on their control counterparts.
    }
    \label{model target ProPres}
\end{figure*}

The two models do not mirror humans in performance on the modal, either.
Their majority labels in the modal with affirmative and negative hypotheses (e.g., $P_5$ with $H_1$ and $H_2$) are entailment and contradiction, respectively.
These results suggest that in the modal, they rely on the lexical overlap heuristic if a hypothesis is affirmative but they adopt a negation heuristic if it is negative, overriding the lexical overlap heuristic.
Specifically, they label a hypothesis as entailment if it is affirmative whereas if \textit{not} is present in it, they label it as contradiction. 

\begin{figure}[t]
    \centering
    \includegraphics[trim=0 0 0 0.1cm,clip, width=\linewidth]{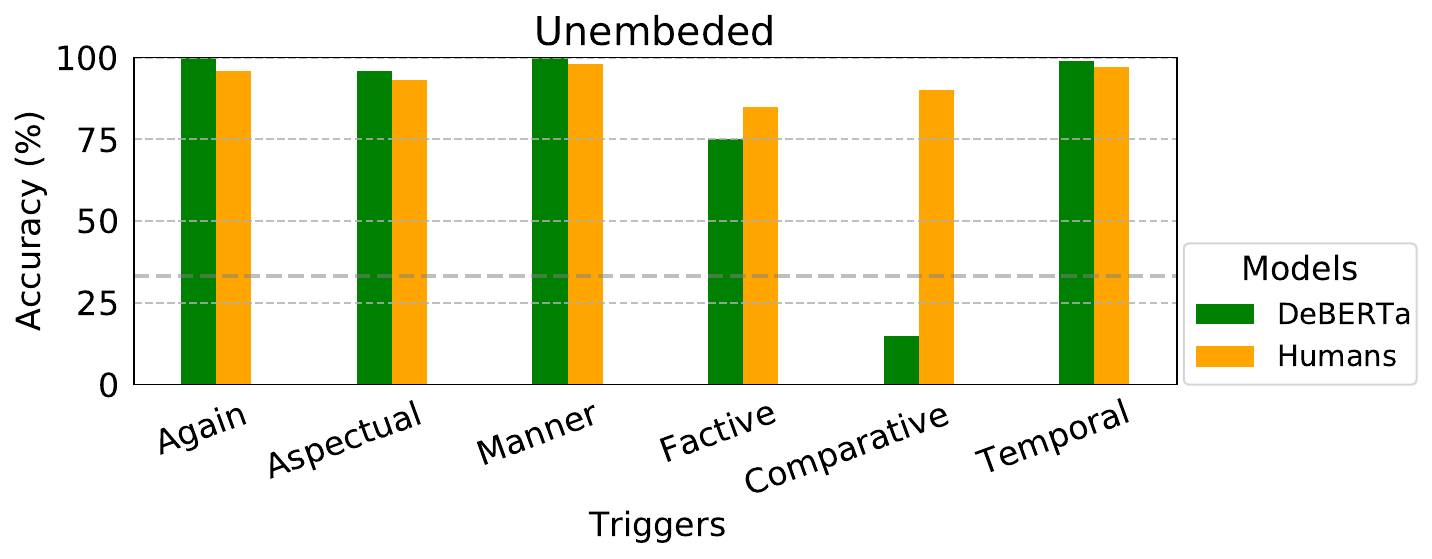} % , height=3cm
    \caption{Results on the unembedded condition in \textsc{ProPres} for DeBERTa and humans.
    }
    \label{ProPres unembedded}
\end{figure}

These variable results for DeBERTa and RoBERTa are inconsistent with \citet{jeretic-etal-2020-natural}, who find that BERT achieves high accuracy for the interrogative and modal controls by correctly assigning them the neutral label.
The discrepancy between our results and \citet{jeretic-etal-2020-natural}'s indicates that the combination of the two environments with new triggers in \textsc{ProPres} makes a more thorough model evaluation possible.

Overall, the performance of RoBERTa and DeBERTa is interpretable regarding the three environments: unembedded, negation, and conditional; hence, we omit model results on the interrogative and modal below.\footnote{We report all results including excluded conditions in Appendix~\ref{appendix results without exclusion}.}
Additionally, since the two models are comparable in accuracy, we only report DeBERTa's performance in what follows.

\paragraph{Unembedded Triggers}
Figure~\ref{ProPres unembedded} shows results on the unembedded triggers.
Overall, DeBERTa and humans achieve high accuracy for all triggers.
One exception is DeBERTa's poor performance on the comparative (e.g., \textit{the girl read the letter better than the boy $\rightarrow$ the boy read the letter}) (14.5\%), indicating its limited knowledge of this trigger.
Hence, we exclude DeBERTa's predictions about the comparative when we report results on entailment-canceling environments.

\paragraph{Entailment-Canceling Environments}
Figure~\ref{model target ProPres} shows results on the entailment-canceling environments.
Our human results provide evidence for variable projectivity (range 55.1--99.8\%).

First, the human results indicate that the iterative \textit{again} weakly projects over the negation (75.8\%) compared to the other three environments (86.3\% on average).
We provide the example sentence pairs for \textit{again} embedded under negation below.

\begin{exe}
\ex 
\label{again in neg}
    $P$: The man did not shed tears again.\\
    $H_{1(2)}$: The man had (not) shed tears before.
\end{exe}

\noindent We reason that this apparent low projectivity is attributable to the fact that the negative sentence with \textit{again} is ambiguous as to whether \textit{again} takes scope over the proposition with negation or without negation \cite{bale2007quantifiers}.
In the first reading, the presupposition is that the man had shed tears before; in the second reading, it is that the man had not shed tears before.
If humans infer the second presupposition, they should label the hypotheses such as $H_{1}$ and $H_{2}$ as entailment and contradiction, respectively, giving rise to the seemingly low projectivity rates.
Since this ambiguity itself has nothing to do with the projectivity, we leave it open whether the observed rate (75.8\%) truly reflects the projectivity or not.
Contrary to humans, the DeBERTa judges the same condition as projective (95\%), indicating that it virtually always predicts the second presupposition (e.g., the man had shed tears before).

Next, manner adverbs exhibit relatively weak projectivity over the negation (e.g., $P_1$ in (\ref{manner examples in text})) and interrogative (e.g., $P_2$) (58.3\% and 66.6\%, respectively).

\begin{exe}
\ex 
\label{manner examples in text}
    $P_1$: The man did not hurt others seriously.\\
    $P_2$: Did the man hurt others seriously?\\
    $P_3$: If the man had hurt others seriously, ... \\
    $P_4$: The man might hurt others seriously.\\ 
    $H_{1(2)}$: The man (did not) hurt others.
\end{exe}

\noindent According to \citet{stevens2017rational} and \citet{tonhauser2019information}, a focalized element in the utterance affects the projectivity of the presupposition introduced by manner adverbs in interrogatives and negation.
For instance, the presupposition ($H_1$) is more likely to project when the focus falls into the manner adverb (\textit{did the man hurt others SERIOUSLY?}) than when it falls into the subject (\textit{did the MAN hurt others seriously?}).
Since our human evaluation provides no prosodic information signaling focus, humans might find these conditions ambiguous, yielding weak projectivity.
Furthermore, our item-by-item analysis with human data reveals that in the manner adverbs embedded under negation, the projectivity ranges between 43.3\% (for \textit{angrily}) and 66.6\% (for \textit{easily}), indicating the within-trigger-type variability.

Adding to \citet{stevens2017rational} and \citet{tonhauser2019information}, we find that the manner adverbs are weakly projective in the conditional (e.g., $P_3$) and modal (e.g., $P_4$) (62.0\% and 55.1\%, respectively).
This suggests that information structural cues such as prosodic focus play a role in the projectivity of presupposition introduced by the manner adverbs embedded under the conditional and modal.

Third, in the modal, temporal adverbs (e.g., $P_1$ in (\ref{temporal adverb examples in text})) and comparatives (e.g., $P_2$) have weaker projectivity (54.7\% and 57.4\%, respectively) than the other three triggers excluding the manner adverbs (92.5\% on average).
These two triggers are projective in the other three environments (79.7\% and 93.4\% on average for the temporal adverbs and comparatives, respectively).
This indicates that the projectivity of presuppositions of these triggers varies depending on the environment.

\begin{exe}
\ex 
\label{temporal adverb examples in text}
    $P_1$: Tom might sing after reading.\\
    $P_2$: The lady might sing better than Tom.\\
    $H_\textit{1(2)}$: Tom (did not) read.
\end{exe}

DeBERTa's performance does not mirror humans' in some cases.
It predicts that the manner adverbs in the negation and conditional ($P_1$ and $P_3$ in (\ref{manner examples in text}), respectively) are not projective (8.5\% and 14\%, respectively), contrary to humans (58.3\% and 62.0\%, respectively).
This indicates that either DeBERTa lacks the knowledge of these two cases or processes them as if the subject is focalized (e.g., \textit{did the MAN hurt others seriously?}).
%DeBERTa takes the other six conditions as projective (range 71.5--99.5\%), similar to humans.

In summary, the human evaluation in Experiment 2 shows variable projectivity in six out of the 24 new conditions, contrary to the first one, in which we observe it in two out of 24 conditions.
This contrast highlights that the combination of various triggers and environments can lead to the discovery of new cases of variable projectivity.
In addition, we find that DeBERTa does not capture variable projectivity in some cases, suggesting that DeBERTa's ability to process presupposition is not necessarily human-like.

\section{Conclusion}
Our experiments reveal that 
humans exhibit the variable projectivity of presupposition in some conditions (two out of 24 and six out of 24 conditions in Experiments 1 and 2, respectively), but the best-performed model, DeBERTa, does not capture it most of the time, indicating that it does not generalize pragmatic inferences for presuppositions.

In our experiments, quite a few conditions are excluded from the analysis for various reasons such as lexical ambiguity in some items, disagreements in human judgments, and the models' lack of knowledge.
To tease apart these factors carries us well beyond the scope of this study.
However, this fact suggests that we need to be careful with dataset creation so that we can train or evaluate models on well-designed datasets targeting pragmatic inferences.

\section*{Acknowledgments}
We would like to thank the anonymous reviewers for their helpful comments.
This work was supported by JST PRESTO Grant Number JPMJPR20C4 and JSPS KAKENHI Grant Number 22K17954.

\bibliography{anthology,custom}
\bibliographystyle{acl_natbib}

% \newpage
\appendix

\section{Limitations}
One of the limitations of our study is that not all data have human labels.
However, it is not feasible to get many judgments for all the data in \textsc{ImpPres} and \textsc{ProPres} in terms of cost.
Extending this study, we hope to conduct a targeted human evaluation with some of the triggers that exhibit the variable projectivity (e.g., manner adverbs).

The second limitation has to do with humans' low accuracy in control modal and question conditions.
We attribute this to the procedure of our evaluation.
The participants are asked to judge whether the hypothesis contradicts, entails, or is neutral to the question or modal premise.
Since it is hard to imagine the situation in which the modal and question sentences are true or false, people might be confused with the instruction.
We hope to collect more valid data using a better instruction in our future study.

The third limitation is that we do not conduct the thorough analyses of between-item variability and between-participant variability in data from the two human evaluations.
It is likely that the projectivity of the presupposition depends on lexical items and participants.
We take these into consideration in the future study.

The final limitation is that this study investigates presuppositions without any context.
Taking \textit{John did not stop cutting trees} as an example, whether the presupposition \textit{John had cut trees before} projects over negation depends on a context.
For instance, the presupposition does not project over negation if we associate the sentence with the appropriate context.
Consider the following example: \textit{Mary liked cutting trees but never smoked. In contrast, John never cut trees but liked smoking. One day Mary and John stopped cutting trees and smoking, respectively.
Later Bob said to Ken ``John stopped cutting trees.''
Then Ken responded ``wait, John didn't stop cutting trees but he stopped smoking''}.
In this example, the sentence \textit{John did not stop cutting trees} does not presuppose \textit{John had cut trees before}.
It remains to be seen how the contextual information affects each trigger embedded under different environments.

\section{Templates} \label{appendix templates}
Tables~\ref{active templates}--\ref{modal templates} contain templates of premises and hypotheses for six triggers crossed with five environments in \textsc{ProPres}.

\begin{table*}
\small \centering
\begin{tabular}{ccc} \toprule
Trigger & Template & Premise and Hypothesis \\ \midrule
\begin{tabular}{c}
\textit{Again}\\
\end{tabular}
& 
\begin{tabular}{c}
$P$: The N VP again.\\
$H_1$: The N had VP before.\\
  
$H_2$: The N had not VP before.\\
\end{tabular}
&
\begin{tabular}{c}
$P$: The doctor shed tears again.\\
 
$H_1$: The doctor had cut the tree before.\\
  
$H_2$: The doctor had not shed tears before.\\
\end{tabular}
\\
\\
\begin{tabular}{c}
Manner \\
adverbs\\
\end{tabular}
& 
\begin{tabular}{c}
$P$: The N VP MADV.\\
 
$H_1$: The N VP.\\
  
$H_2$: The N did not VP.\\
\end{tabular}
&
\begin{tabular}{c}
$P$: The doctor shed tears slowly.\\
 
$H_1$: The doctor shed tears.\\
  
$H_2$: The doctor did not shed tears.\\
\end{tabular}
\\
\\
\begin{tabular}{c}
Comparatives\\
\end{tabular}
& 
\begin{tabular}{c}
$P$: The $\text{N}_1$ VP ADVer than $\text{N}_2$.\\
 
$H_1$: The $\text{N}_2$ VP.\\
  
$H_2$: The $\text{N}_2$ did not VP.\\
\end{tabular}
&
\begin{tabular}{c}
$P$: The doctor shed tears better than the singer.\\
 
$H_1$: The singer shed tears.\\
  
$H_2$: The singer did not shed tears.\\
\end{tabular}
\\
\\
\begin{tabular}{c}
Temporal \\
adverbs\\
\end{tabular}
& 
\begin{tabular}{c}
$P$: The N $\text{VP}_1$ TADV $\text{VP}_2$ing.\\
 
$H_1$: The N $\text{VP}_2$.\\
  
$H_2$: The N did not $\text{VP}_2$.\\
\end{tabular}
&
\begin{tabular}{c}
$P$: The doctor shed tears before hurting others.\\
 
$H_1$: The doctor hurt others.\\
  
$H_2$: The doctor did not hurt others\\
\end{tabular}
\\
\\
\begin{tabular}{c}
Aspectual \\
verbs\\
\end{tabular}
& 
\begin{tabular}{c}
$P$: The N ASP VPing.\\
 
$H_1$: The N had been VPing.\\
  
$H_2$: The N had not been VPing.\\
\end{tabular}
&
\begin{tabular}{c}
$P$: The doctor stopped shedding tears.\\
 
$H_1$: The doctor had been shedding tears.\\
  
$H_2$: The doctor had not been shedding tears.\\
\end{tabular}
\\
\\
\begin{tabular}{c}
Factive \\
verbs\\
\end{tabular}
& 
\begin{tabular}{c}
$P$: The N Factive VPing.\\
 
$H_1$: The N VP.\\
  
$H_2$: The N did not VP.\\
\end{tabular}
&
\begin{tabular}{c}
$P$: The doctor regretted shedding tears.\\
 
$H_1$: The doctor shed tears.\\
  
$H_2$: The doctor shed tears.\\
\end{tabular}
\\ \bottomrule
\end{tabular}
\caption{Templates for affirmative sentences.}
\label{active templates}
\end{table*}

\begin{table*}
\small \centering
\setlength\tabcolsep{4pt}
\begin{tabular}{ccc} \toprule
Trigger & Template & Premise and Hypothesis \\ \midrule
\begin{tabular}{c}
\textit{Again}\\
\end{tabular}
& 
\begin{tabular}{c}
$P$: The N did not VP again.\\
 
$H_1$: The N had VP before.\\
  
$H_2$: The N had not VP before.\\
\end{tabular}
&
\begin{tabular}{c}
$P$: The doctor did not shed tears again.\\
 
$H_1$: The doctor had shed tears before.\\
  
$H_2$: The doctor had not shed tears before.\\
\end{tabular}
\\
\\
\begin{tabular}{c}
Manner \\
adverbs\\
\end{tabular}
& 
\begin{tabular}{c}
$P$: The N did not VP MADV.\\
 
$H_1$: The N VP.\\
  
$H_2$: The N did not VP.\\
\end{tabular}
&
\begin{tabular}{c}
$P$: The doctor did not shed tears slowly.\\
 
$H_1$: The doctor shed tears.\\
  
$H_2$: The doctor did not shed tears.\\
\end{tabular}
\\
\\
\begin{tabular}{c}
Comparatives\\
\end{tabular}
& 
\begin{tabular}{c}
$P$: The $\text{N}_1$ did not VP ADVer than $\text{N}_2$.\\
 
$H_1$: The $\text{N}_2$ VP.\\
  
$H_2$: The $\text{N}_2$ did not VP.\\
\end{tabular}
&
\begin{tabular}{c}
$P$: The doctor did not shed tears better than the singer.\\
 
$H_1$: The singer shed tears.\\
  
$H_2$: The singer did not shed tears.\\
\end{tabular}
\\
\\
\begin{tabular}{c}
Temporal \\
adverbs\\
\end{tabular}
& 
\begin{tabular}{c}
$P$: The N did not $\text{VP}_1$ TADV $\text{VP}_2$ing.\\
 
$H_1$: The N $\text{VP}_2$.\\
  
$H_2$: The N did not $\text{VP}_2$.\\
\end{tabular}
&
\begin{tabular}{c}
$P$: The doctor did not shed tears before hurting others.\\
 
$H_1$: The doctor hurt others.\\
  
$H_2$: The doctor did not hurt others.\\
\end{tabular}
\\
\\
\begin{tabular}{c}
Aspectual \\
verbs\\
\end{tabular}
& 
\begin{tabular}{c}
$P$: The N did not ASP VPing.\\
 
$H_1$: The N had been VPing.\\
  
$H_2$: The N had not been VPing.\\
\end{tabular}
&
\begin{tabular}{c}
$P$: The doctor did not stop shedding tears.\\
 
$H_1$: The doctor had been shedding tears.\\
  
$H_2$: The doctor had not been shedding tears.\\
\end{tabular}
\\
\\
\begin{tabular}{c}
Factive\\
verbs\\
\end{tabular}
& 
\begin{tabular}{c}
$P$: The N did not Factive VPing.\\
 
$H_1$: The N VP.\\
  
$H_2$: The N did not VP.\\
\end{tabular}
&
\begin{tabular}{c}
$P$: The doctor did not regret shedding tears.\\
 
$H_1$: The doctor shed tears.\\
  
$H_2$: The doctor did not shed tears.\\
\end{tabular}
\\ \bottomrule
\end{tabular}
\caption{Templates for negative sentences.}
\label{negation templates}
\end{table*}

\begin{table*}
\small \centering
\setlength\tabcolsep{4pt}
\begin{tabular}{ccc} \toprule
Trigger & Template & Premise and Hypothesis \\ \midrule
\begin{tabular}{c}
\textit{Again}\\
\end{tabular}
& 
\begin{tabular}{c}
$P$: Did the N VP again?\\
 
$H_1$: The N had VP before.\\
  
$H_2$: The N had not VP before.\\
\end{tabular}
&
\begin{tabular}{c}
$P$: Did the doctor shed tears again?\\
 
$H_1$: The doctor had shed tears before.\\
  
$H_2$: The doctor had not shed tears before.\\
\end{tabular}
\\
\\
\begin{tabular}{c}
Manner \\
adverbs\\
\end{tabular}
& 
\begin{tabular}{c}
$P$: Did the N VP MADV?\\
 
$H_1$: The N VP.\\
  
$H_2$: The N did not VP.\\
\end{tabular}
&
\begin{tabular}{c}
$P$: Did the doctor shed tears slowly?\\
 
$H_1$: The doctor shed tear.\\
  
$H_2$: The doctor did not shed tears.\\
\end{tabular}
\\
\\
\begin{tabular}{c}
Comparatives\\
\end{tabular}
& 
\begin{tabular}{c}
$P$: Did the $\text{N}_1$ VP ADVer than $\text{N}_2$?\\
 
$H_1$: The $\text{N}_2$ VP.\\
  
$H_2$: The $\text{N}_2$ did not VP.\\
\end{tabular}
&
\begin{tabular}{c}
$P$: Did the doctor shed tears better than the singer?\\
 
$H_1$: The doctor shed tears.\\
  
$H_2$: The doctor did not shed tears.\\
\end{tabular}
\\
\\
\begin{tabular}{c}
Temporal \\
adverbs\\
\end{tabular}
& 
\begin{tabular}{c}
$P$: Did the N $\text{VP}_1$ TADV $\text{VP}_2$ing?\\
 
$H_1$: The N $\text{VP}_2$.\\
  
$H_2$: The N did not $\text{VP}_2$.\\
\end{tabular}
&
\begin{tabular}{c}
$P$: Did the doctor shed tears before spreading the rumor?\\
 
$H_1$: The doctor spread the rumor.\\
  
$H_2$: The doctor did not spread the rumor.\\
\end{tabular}
\\
\\
\begin{tabular}{c}
Aspectual \\
verbs\\
\end{tabular}
& 
\begin{tabular}{c}
$P$: Did the N ASP VPing?\\
 
$H_1$: The N had been VPing.\\
  
$H_2$: The N had not been VPing.\\
\end{tabular}
&
\begin{tabular}{c}
$P$: Did the doctor stop shedding tears?\\
 
$H_1$: The doctor had been shedding tears.\\
  
$H_2$: The doctor had not been shedding tears.\\
\end{tabular}
\\
\\
\begin{tabular}{c}
Factive \\
verbs\\
\end{tabular}
& 
\begin{tabular}{c}
$P$: Did the N Factive VPing?\\
 
$H_1$: The N VP.\\
  
$H_2$: The N did not VP.\\
\end{tabular}
&
\begin{tabular}{c}
$P$: Did the doctor regret shedding tears?\\
 
$H_1$: The doctor shed tears.\\
  
$H_2$: The doctor did not shed tears.\\
\end{tabular}
\\ \bottomrule
\end{tabular}
\caption{Templates for interrogatives.}
\label{question templates}
\end{table*}

\begin{table*}
\small \centering
\setlength\tabcolsep{4pt}
\begin{tabular}{ccc} \toprule
Trigger & Template & Examples \\ \midrule
\begin{tabular}{c}
\textit{Again}\\
\end{tabular}
& 
\begin{tabular}{c}
$P$: If the $\text{N}_1$ had VP again,\\
the $\text{N}_2$ would have $\text{VP}_2$.\\
 
$H_1$: The $\text{N}_1$ had $\text{VP}_1$ before.\\
  
$H_2$: The $\text{N}_1$ had not $\text{VP}_1$ before.\\
\end{tabular}
&
\begin{tabular}{c}
$P$: If the doctor had shed tears again,\\
the singer could have spread the news.\\
 
$H_1$: The doctor had shed tears before.\\
  
$H_2$: The doctor had not shed tears before.\\
\end{tabular}
\\
\\
\begin{tabular}{c}
Manner \\
adverbs\\
\end{tabular}
& 
\begin{tabular}{c}
$P$: If the $\text{N}_1$ $\text{VP}_1$ MADV,\\
the $\text{N}_2$ would have $\text{VP}_2$.\\
 
$H_1$: The $\text{N}_1$ $\text{VP}_1$.\\
  
$H_2$: The $\text{N}_1$ did not $\text{VP}_1$.\\
\end{tabular}
&
\begin{tabular}{c}
$P$: If the doctor shed tears slowly,\\
the singer could have spread the news.\\
 
$H_1$: The doctor shed tears.\\
  
$H_2$: The doctor did not shed tears.\\
\end{tabular}
\\
\\
\begin{tabular}{c}
Comparatives\\
\end{tabular}
& 
\begin{tabular}{c}
$P$: If the $\text{N}_1$ had $\text{VP}_1$ ADVer than \\
$\text{N}_3$, the $\text{N}_2$ would have $\text{VP}_2$.\\
 
$H_1$: The $\text{N}_1$ $\text{VP}_1$.\\
  
$H_2$: The $\text{N}_1$ did not $\text{VP}_1$.\\
\end{tabular}
&
\begin{tabular}{c}
$P$: If the doctor had shed tears better than the singer,\\
the artist could have spread the news.\\
 
$H_1$: The singer shed tears.\\
  
$H_2$: The singer did not shed tears.\\
\end{tabular}
\\
\\
\begin{tabular}{c}
Temporal \\
adverbs\\
\end{tabular}
& 
\begin{tabular}{c}
$P$: If the $\text{N}_1$ had $\text{VP}_1$ TADV $\text{VP}_2$ing,\\
the $\text{N}_2$ would have $\text{VP}_3$.\\
 
$H_1$: The $\text{N}_1$ $\text{VP}_2$.\\
  
$H_2$: The $\text{N}_1$ did not $\text{VP}_2$.\\
\end{tabular}
&
\begin{tabular}{c}
$P$: If the doctor had shed tears before spreading the rumor,\\
the singer could have burst into the room.\\
 
$H_1$: The doctor spread the rumor.\\
  
$H_2$: The doctor did not spread the rumor.\\
\end{tabular}
\\
\\
\begin{tabular}{c}
Aspectual \\
verbs\\
\end{tabular}
& 
\begin{tabular}{c}
$P$: If the $\text{N}_1$ ASP $\text{VP}_1$ing,\\
the $\text{N}_2$ would have $\text{VP}_2$.\\
 
$H_1$: The $\text{N}_1$ had been $\text{VP}_1$ing.\\
  
$H_2$: The $\text{N}_1$ had not been $\text{VP}_1$ing.\\
\end{tabular}
&
\begin{tabular}{c}
$P$: If the doctor had stopped shedding tears,\\
the singer could have spread the rumor.\\
 
$H_1$: The doctor had been shedding tears.\\
  
$H_2$: The doctor had not been shedding tears.\\
\end{tabular}
\\
\\
\begin{tabular}{c}
Factive \\
verbs\\
\end{tabular}
& 
\begin{tabular}{c}
$P$: If the $\text{N}_1$ Factive $\text{VP}_1$ing,\\
the $\text{N}_2$ would have $\text{VP}_2$.\\
 
$H_1$: The $\text{N}_1$ $\text{VP}_1$.\\
  
$H_2$: The $\text{N}_1$ did not $\text{VP}_1$.\\
\end{tabular}
&
\begin{tabular}{c}
$P$: If the doctor had regretted shedding tears,\\
the singer could have spread the rumor.\\
 
$H_1$: The doctor shed tears.\\
  
$H_2$: The doctor did not shed tears.\\
\end{tabular}
\\ \bottomrule
\end{tabular}
\caption{Templates for counterfactual conditionals.}
\label{conditional templates}
\end{table*}

\begin{table*}
\small \centering
\setlength\tabcolsep{4pt}
\begin{tabular}{ccc} \toprule
Trigger & Template & Premise and Hypothesis \\ \midrule
\begin{tabular}{c}
Again\\
\end{tabular}
& 
\begin{tabular}{c}
$P$: The N Modal VP again.\\
 
$H_1$: The N had VP before.\\
  
$H_2$: The N had not VP before.\\
\end{tabular}
&
\begin{tabular}{c}
$P$: The doctor might shed tears again.\\
 
$H_1$: The doctor had shed tears before.\\
  
$H_2$: The doctor had not shed tears before.\\
\end{tabular}
\\
\\
\begin{tabular}{c}
Manner \\
adverbs\\
\end{tabular}
& 
\begin{tabular}{c}
$P$: The N Modal VP MADV.\\
 
$H_1$: The N VP.\\
  
$H_2$: The N did not VP.\\
\end{tabular}
&
\begin{tabular}{c}
$P$: The doctor might shed tears slowly.\\
 
$H_1$: The doctor shed tears.\\
  
$H_2$: The doctor did not shed tears.\\
\end{tabular}
\\
\\
\begin{tabular}{c}
Comparatives\\
\end{tabular}
& 
\begin{tabular}{c}
$P$: The $\text{N}_1$ Modal VP ADVer than $\text{N}_2$.\\
 
$H_1$: The $\text{N}_2$ VP.\\
  
$H_2$: The $\text{N}_2$ did not VP.\\
\end{tabular}
&
\begin{tabular}{c}
$P$: The doctor might shed tears better than the singer.\\
 
$H_1$: The singer shed tears.\\
  
$H_2$: The singer did not shed tears.\\
\end{tabular}
\\
\\
\begin{tabular}{c}
Temporal \\
adverbs\\
\end{tabular}
& 
\begin{tabular}{c}
$P$: The N Modal $\text{VP}_1$ TADV $\text{VP}_2$ing.\\
 
$H_1$: The N $\text{VP}_2$.\\
  
$H_2$: The N did not $\text{VP}_2$.\\
\end{tabular}
&
\begin{tabular}{c}
$P$: The man might shed tears before spreading the rumor.\\
 
$H_1$: The man spread the rumor.\\
  
$H_2$: The man did not spread the rumor.\\
\end{tabular}
\\
\\
\begin{tabular}{c}
Aspectual\\
verbs\\
\end{tabular}
& 
\begin{tabular}{c}
$P$: The N Modal ASP VPing.\\
 
$H_1$: The N had been VPing.\\
  
$H_2$: The N had not been VPing.\\
\end{tabular}
&
\begin{tabular}{c}
$P$: The doctor might stop shedding tears.\\
 
$H_1$: The doctor had been shedding tears.\\
  
$H_2$: The doctor had not been shedding tears.\\
\end{tabular}
\\
\\
\begin{tabular}{c}
Factive\\
verbs\\
\end{tabular}
& 
\begin{tabular}{c}
$P$: The N Modal Factive VPing.\\
 
$H_1$: The N VP.\\
  
$H_2$: The N did not VP.\\
\end{tabular}
&
\begin{tabular}{c}
$P$: The doctor might regret shedding tears.\\
$H_1$: The doctor shed tears.\\
  
$H_2$: The doctor did not shed tears.\\
\end{tabular}
\\ \bottomrule
\end{tabular}
\caption{Templates for modal sentences.}
\label{modal templates}
\end{table*}

\section{Crowdsourcing Human Evaluation} \label{qualification}
Before the experiment, each participant is asked to read a written instruction about the NLI task carefully.
All data are collected anonymously except workers' ID.

\paragraph{Evaluation 1}
\begin{figure}[t]
    \centering
    \includegraphics[trim=0 0 0 0.1cm,clip, width=\linewidth]{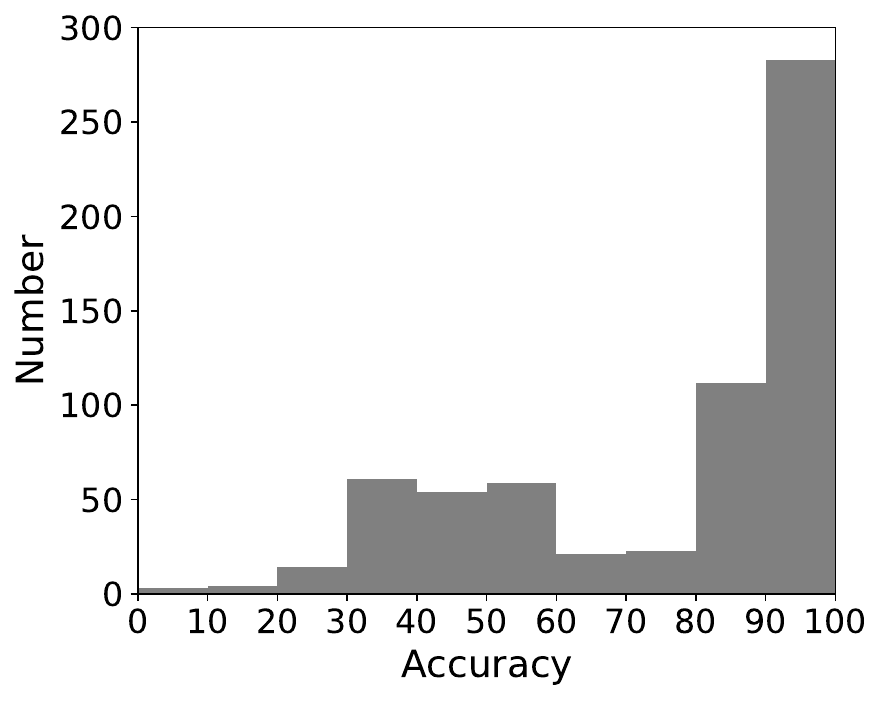}  % , height=4cm
    \caption{Distributions of accuracy in the control conditions in \textsc{ProPres}.}
    \label{control accuracy histogram}
\end{figure}

\noindent Using Amazon Mechanical Turk, we recruit 116 people with the requirements of having an approval rating of 99.0\% or higher, having at least 5,000 approved tasks, being
located in the US, the UK, or Canada, and having passed a qualification task.
We make sure that the workers are paid at least \$12.0 USD per hour.
Among them, we exclude the responses of 46 participants from the analysis because their accuracy rates for a sanity check are below 80.0\%.
We analyze the data of the remaining 71 participants.

\paragraph{Evaluation 2}
Using Amazon Mechanical Turk, we recruit 635 people with the requirements of having an approval rating of 99.0\% or
higher, having at least 5,000 approved tasks, and being
located in the US, the UK, or Canada.
We make sure that the workers are paid at least \$12.0 USD per hour.
Among them, we exclude the responses of 352 participants whose accuracy for the control conditions is less than 90\% based on the distributions of accuracy in Figure~\ref{control accuracy histogram}.
The control results include results for unembedded, negation, and conditional conditions.
The interrogative control condition is not included in the mean calculation, because its mean accuracy is around chance (36.0\% over the chance level 33.3\%).
As a result, we analyze the data of the remaining 283 participants.

\section{Triggers and Environments in \textsc{ImpPres}}
\label{appendix triggers and environments in imppres}
Tables~\ref{triggers in imppres} and \ref{environments in imppres} present triggers and environments used in \textsc{ImpPres}, respectively.

\begin{table*}
\small
\centering \def\arraystretch{1.2}
\setlength{\tabcolsep}{4pt}
\rowcolors{1}{white}{gray!25}
\begin{tabular}{ccc} \toprule
Trigger & Example & Presupposition \\ \midrule
\begin{tabular}{c}
\textit{All N}
\end{tabular}
& 
\begin{tabular}{c}
All four waiters that bothered Paul telephoned.
\end{tabular}
& 
\begin{tabular}{c}
Exactly four waiters telephoned.
\end{tabular}
\\
\begin{tabular}{c}
\textit{Both}
\end{tabular}
& 
\begin{tabular}{c}
Both people that hoped to move have married.
\end{tabular}
& 
\begin{tabular}{c}
Exactly two people have married.
\end{tabular}
\\
\begin{tabular}{c}
Change of state verb
\end{tabular}
& 
\begin{tabular}{c}
Marie was leaving.
\end{tabular}
& 
\begin{tabular}{c}
Marie was here.
\end{tabular}
\\
\begin{tabular}{c}
Cleft existence
\end{tabular}
& 
\begin{tabular}{c}
It is Margaret that forgot Dan.
\end{tabular}
& 
\begin{tabular}{c}
Someone forgot Dan.
\end{tabular}
\\
\begin{tabular}{c}
Cleft uniqueness
\end{tabular}
& 
\begin{tabular}{c}
It is Donna who studied.
\end{tabular}
& 
\begin{tabular}{c}
Exactly one person studied.
\end{tabular}
\\
\begin{tabular}{c}
\textit{Only}
\end{tabular}
& 
\begin{tabular}{c}
The pasta only annoys Roger.
\end{tabular}
& 
\begin{tabular}{c}
The pasta annoys Roger.
\end{tabular}
\\
\begin{tabular}{c}
Possessive definites
\end{tabular}
& 
\begin{tabular}{c}
The boy's rugs did look like these prints.
\end{tabular}
& 
\begin{tabular}{c}
The boy has rugs.
\end{tabular}
\\
\begin{tabular}{c}
Possessive uniqueness
\end{tabular}
& 
\begin{tabular}{c}
Maria's apple that ripened annoys the boy.
\end{tabular}
& 
\begin{tabular}{c}
Maria has exactly one apple that ripened.
\end{tabular}
\\
\begin{tabular}{c}
Question
\end{tabular}
& 
\begin{tabular}{c}
Bob learns how Rachel approaches Melanie.
\end{tabular}
& 
\begin{tabular}{c}
Rachel approaches Melanie.
\end{tabular}
\\ \bottomrule
\end{tabular}
\caption{Examples of triggers in \textsc{ImpPres}.}
\label{triggers in imppres}
\end{table*}

\begin{table*}
\centering
\setlength{\tabcolsep}{5pt}
\begin{tabular}{cc} \toprule
Environment & Example \\ \midrule
\begin{tabular}{c}
Unembedded 
\end{tabular}
& 
\begin{tabular}{c}
All four waiters that bothered Paul telephoned.
\end{tabular}
\\
\begin{tabular}{c}
Negation
\end{tabular}
& 
\begin{tabular}{c}
All four waiters that bothered Paul did not telephone.
\end{tabular}
\\
\begin{tabular}{c}
Interrogative
\end{tabular}
& 
\begin{tabular}{c}
Did all four waiters that bothered Paul telephone?
\end{tabular}
\\
\begin{tabular}{c}
Conditional
\end{tabular}
& 
\begin{tabular}{c}
If all four waiters that bothered Paul telephoned, it's okay.
\end{tabular}
\\
\begin{tabular}{c}
Modal
\end{tabular}
& 
\begin{tabular}{c}
All four waiters that bothered Paul might telephone.
\end{tabular}
\\ \bottomrule
\end{tabular}
\caption{
    Environments used in \textsc{ImpPres}.
    }
\label{environments in imppres}
\end{table*}

\section{Results without Exclusion}
\label{appendix results without exclusion}
\begin{figure*}[t]
    \centering
    \includegraphics[trim=0 0 0 0.1cm,clip,width=\textwidth]{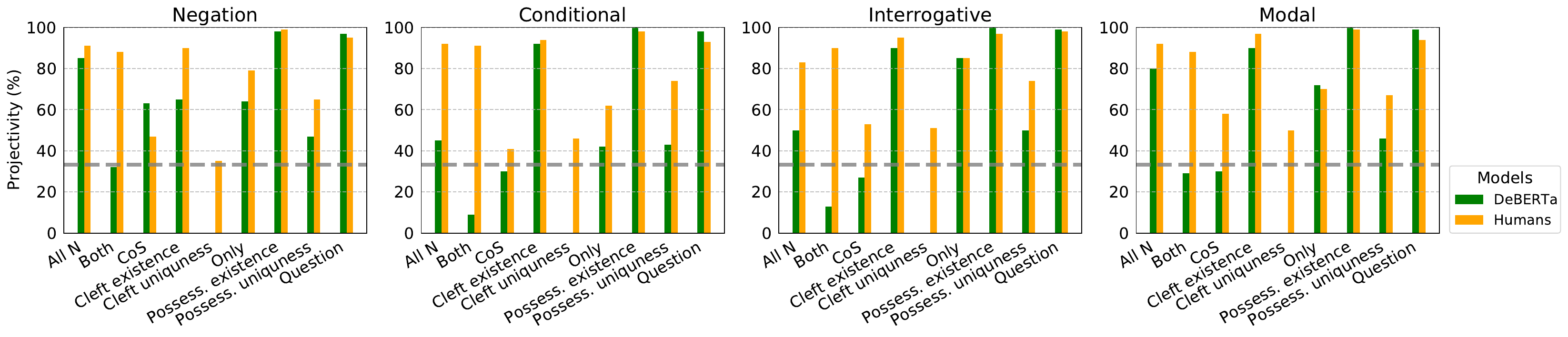} % , height=4cm
    \caption{Results on triggers embedded under the negation, conditional, interrogative, and modal in \textsc{ImpPres}.
    }
    \label{results without exclusion in imppres}
\end{figure*}

\begin{figure*}[t]
    \centering
    \includegraphics[trim=0 0 0 0.1cm,clip,width=\textwidth]{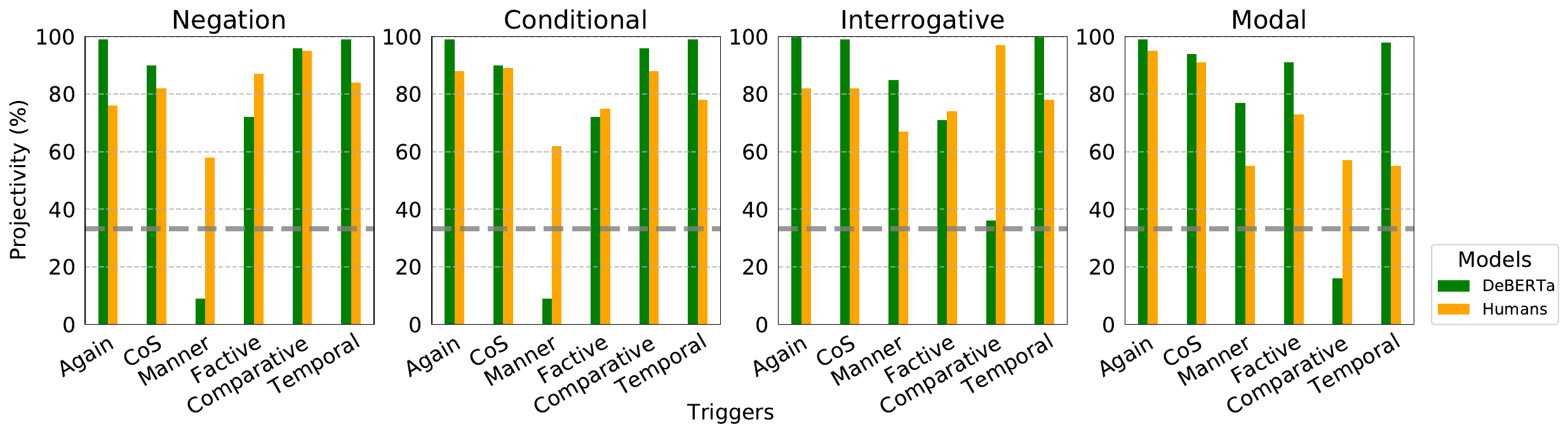} % , height=4cm
    \caption{Results on triggers embedded under the negation, conditional, interrogative, and modal in \textsc{ProPres}.
    }
    \label{results without exclusion in ProPres}
\end{figure*}

Figures~\ref{results without exclusion in imppres} and \ref{results without exclusion in ProPres} present results without exclusion of triggers and environments in \textsc{ImpPres} and \textsc{ProPres}, respectively.

\end{document}